\documentclass{article} 
\usepackage{iclr2021_conference,times}
\usepackage{graphicx}
\iclrfinalcopy

\usepackage{amsmath,amsfonts,bm}

\def\eqref#1{equation~\ref{#1}}

\def\1{\bm{1}}

\DeclareMathAlphabet{\mathsfit}{\encodingdefault}{\sfdefault}{m}{sl}
\SetMathAlphabet{\mathsfit}{bold}{\encodingdefault}{\sfdefault}{bx}{n}

\DeclareMathOperator*{\argmax}{arg\,max}
\DeclareMathOperator*{\argmin}{arg\,min}

\usepackage{amsmath}
\usepackage{amsfonts}
\usepackage{amssymb}
\usepackage{booktabs}

\newlength{\spc} 

\newcommand{\states}{\mathcal{S}}
\newcommand{\actions}{\mathcal{A}}
\newcommand{\obs}{\mathcal{O}}

\usepackage[mathscr]{euscript}

\newcommand{\EE}{\mathbb{E}}

\newcommand{\tss}[1]{\textrm{#1}}

\usepackage{hyperref}
\usepackage{subcaption}
\usepackage{url}
\usepackage{bbm}

\usepackage{xcolor}
\newif\ifshowtopics
\showtopicsfalse      

\title{InfraLib: Enabling Reinforcement Learning and Decision-Making for Large-Scale Infrastructure Management}

\author{Pranay Thangeda \& Melkior Ornik \\
Department of Aerospace Engineering\\
University of Illinois Urbana-Champaign\\
\texttt{\{pranayt2,mornik\}@illinois.edu} \\
\AND
Trevor S. Betz \& Michael N. Grussing \\
Engineer Research and Development Center \\
US Army Corps of Engineers \\
\texttt{\{trevor.s.betz,michael.n.grussing\}@usace.army.mil}
}

\begin{document}

\maketitle

\begin{abstract}
Efficient management of infrastructure systems is crucial for economic stability, sustainability, and public safety. However, infrastructure sustainment is challenging due to the vast scale of systems, stochastic deterioration of components, partial observability, and resource constraints. Decision-making strategies that rely solely on human judgment often result in suboptimal decisions over large scales and long horizons. While data-driven approaches like reinforcement learning offer promising solutions, their application has been limited by the lack of suitable simulation environments. We present InfraLib, an open-source modular and extensible framework that enables modeling and analyzing infrastructure management problems with resource constraints as sequential decision-making problems. The framework implements hierarchical, stochastic deterioration models, supports realistic partial observability, and handles practical constraints including cyclical budgets and component unavailability. InfraLib provides standardized environments for benchmarking decision-making approaches, along with tools for expert data collection and policy evaluation. Through case studies on both synthetic benchmarks and real-world road networks, we demonstrate InfraLib's ability to model diverse infrastructure management scenarios while maintaining computational efficiency at scale.
\end{abstract}

\section{Introduction}
Infrastructure systems are the backbone of modern society, encompassing a wide array of essential services including transportation networks, utility systems, and public facilities. Efficient infrastructure management is crucial for modern society's functioning, influencing economic stability \cite{egert2009infrastructure,srinivasu2013infrastructure}, environmental sustainability \cite{yang2018towards}, and public safety \cite{zimmerman2001social}. Managing modern infrastructure systems is a complex and multifaceted task, involving the inspection, maintenance, and replacement of numerous components distributed across facilities and networks \cite{rinaldi2001identifying,kabir2014review}. The challenges of infrastructure management are further compounded by their vast scale, the stochastic nature of component deterioration \cite{madanat1993optimal,frangopol2019bridge}, stringent operational constraints \cite{makovsek2015efficiency}, limited resources \cite{kabir2014review}, and extreme weather events due to climate change \cite{mcdaniels2008fostering,wilbanks2013climate,linkov2014changing}.

Traditional infrastructure sustainment approaches often rely on human expertise, rule-based methodologies, and deterministic models. While these methods can be effective in controlled settings, they struggle to capture the inherent uncertainties and dynamic variations present in real-world scenarios \cite{srinivasu2013infrastructure}. Additionally, they may not adequately address the complexities involved in strategic resource allocation and budgetary considerations, which are critical for effective management \cite{abaza2002optimum,denysiuk2017two}.

In recent years, data-driven methodologies, particularly those leveraging machine learning techniques like reinforcement learning (RL) and imitation learning (IL), have gained attention \cite{hussein2017imitation,sutton2018reinforcement}. These approaches offer promising avenues for decision-making under uncertainty, enabling adaptive and proactive infrastructure management strategies \cite{andriotis2019managing,yao2020deep}. RL, in particular, has achieved remarkable success in domains where decision-making agents can learn optimal policies by interacting with their environment \cite{silver2016mastering,vinyals2019grandmaster}.

Despite their potential, the application of RL and IL in infrastructure management is still in its infancy. A significant barrier is the lack of suitable simulation environments that can accurately model the complexity, scale, and uncertainties of real-world infrastructure systems \cite{andriotis2019managing,vora2023welfare}. Existing tools often fail to represent the stochastic deterioration and partial observability inherent in these systems, limiting the development and evaluation of advanced learning-based strategies.

To address these challenges, we introduce InfraLib, an open-source framework designed for modeling and analyzing large-scale infrastructure management problems. InfraLib provides a realistic and granular representation of infrastructure systems by integrating a hierarchical model that captures the intricate relationships between different components and facilities \cite{rinaldi2001identifying}. Moreover, InfraLib employs a stochastic approach to mimic the real-world uncertainties and partial observability inherent in infrastructure systems \cite{madanat1993optimal,frangopol2019bridge,thangeda2020protrip}, enabling the development and validation of infrastructure management strategies that are robust to the challenges faced in real-world scenarios. InfraLib can utilize the component data from existing sustainment management systems or other life-cycle attribute databases to create realistic environments for decision-making. InfraLib directly supports the modeling of different infrastructure management problems as RL environments, enabling the development and evaluation of learning-based strategies.

InfraLib serves as more than just a simulation tool for learning-based approaches – it provides a comprehensive graphical interface to enhance human decision-making. Through its analysis features, expert data collection capabilities, and policy evaluation tools, users can explore various "what-if" scenarios and develop human-in-the-loop decision-making policies based on expert data. Infralib also offers standardized environments for benchmarking learning-based approaches alongside traditional optimization methods and rule-based strategies to lower the barrier of entry for decision-making research in infrastructure sustainment.

In this paper, we present a detailed overview of the InfraLib framework, highlighting its capabilities and potential applications. We demonstrate the ability of InfraLib to create realistic scenarios for the deployment and evaluation of learning-based approaches, showcasing its ability to model the complexities and challenges encountered in real-world infrastructure systems. Through a series of benchmarks and environments, we illustrate the utility and scalability of InfraLib in facilitating the development and comparison of novel management strategies.

The rest of this paper is organized as follows. The "Preliminaries and Background" section introduces key concepts and background information on infrastructure management, Partially Observable Markov Decision Processes, and data-driven approaches for decision making. The "POMDP Model of Infrastructure" section and the subsequent section formalize the infrastructure management problem as a POMDP and discuss various research challenges that arise in this context. We then present InfraLib in the next section, detailing its structure, component dynamics, and key functionalities including the human interface aspects of InfraLib. Finally, in the "Experiments and Case Studies" section, we showcase example environments and benchmarks to demonstrate InfraLib's utility, versatility, and scalability. 

\section{Preliminaries and Background}
\label{sec:preliminaries}

In this section, we provide background on the considered infrastructure management setting, including the hierarchical nature of infrastructure systems, the metrics used to quantify component health, the dynamics of component deterioration, and the budget constraints that govern infrastructure management. We also provide an overview of Partially Observable Markov Decision Processes (POMDPs) and data-driven approaches that have been successfully applied to sequentialdecision-making under uncertainty. We begin by establishing the notation used throughout the paper. 

Given a finite set $\actions$, $|\actions|$ denotes its cardinality and $\Delta(\actions)$ denotes the set of all probability distributions over the set $\actions$. $\mathbb{N}_0$ denotes the set of natural numbers including 0, i.e., $\mathbb{N}_0 = \{0,1,2, \ldots\}$. 

\subsection{Infrastructure Hierarchy and Management}

Infrastructure management is inherently hierarchical, comprising multiple layers of organization across diverse domains. The base level consists of individual \textit{components}, which are the smallest units of infrastructure elements. These components are typically organized into subsystems based on their functional relationships and interdependencies. For example, in a building context, components might be grouped into subsystems like heating or plumbing, while in a transportation network, components could be grouped into subsystems like pavement sections or bridge elements. These subsystems form part of larger systems (e.g., HVAC systems in buildings or highway corridors in transportation networks), which in turn may belong to facilities or networks. Multiple facilities or networks may collectively form complexes or regional systems, which are managed under an organization. This hierarchical structuring is inherent in real-world infrastructure systems across domains and is crucial for systematic management and decision-making processes.

The condition of each component in the infrastructure is characterized by a \textit{Condition Index (CI)} \cite{grussing2006condition}, a metric that reflects its health status. The CI of a component quantitatively represents the health of a component and deteriorates over time due to environmental factors, wear-and-tear, and in some cases catastrophically due to a manufacturing defect or an external event. This deterioration is typically stochastic, arising from unpredictable environmental interactions and the inherent characteristics of infrastructure components. Moreover, the CI is not always directly observable, necessitating periodic inspections to estimate its current state. These inspections, while essential, incur additional costs. Even when the CI is observed by inspection, the observation is subjective depending on the inspector and the inspection method, and can be noisy.

Management of infrastructure systems at the component level encompasses a wide range of activities, with the common actions including various types of inspection (e.g., visual assessment, sensor-based monitoring, non-destructive testing), maintenance (both preventative and corrective), repair (ranging from minor fixes to major rehabilitation), and replacement. Inspection activities provide estimates of the CI at varying costs and levels of accuracy, replacement involves completely substituting the component at a higher expense, and repair options, which are typically more cost-effective than replacement, aim to improve the CI. These actions, along with their timing and sequencing, are fundamental to maintaining the overall health of the infrastructure system.

Given the stochastic nature of component deterioration and the partial observability of component conditions, we model the environment as a Partially Observable Markov Decision Process (POMDP) \cite{kaelbling1998planning}. POMDPs provide a natural framework for representing scenarios where an agent must make decisions with incomplete information about the true state of the system, while accounting for both the immediate costs and long-term impacts of these decisions.

\subsection{Partially Observable Markov Decision Process}
A discrete-time finite-horizon POMDP $M$ is specified by the tuple $(\states, \actions, \obs, T, Z, R, H)$, where $\states$ denotes a finite set of states, $\actions$ denotes a finite set of actions, and $\obs$ denotes a finite set of observations. The transition probability function is given by $T :  \states\times A \rightarrow \Delta(\states)$, where $\Delta(\states)$ is the space of probability distributions over $\states$. Furthermore, $Z : \obs\times\states\times \actions \rightarrow \Delta(\obs)$ denotes the observation probability function where $\Delta(\obs)$ is analogous to $\Delta(\states)$. Finally, $R : \states \times \actions \rightarrow [-R_\tss{min}, R_\tss{max}]$ denotes the reward function and $H \in \mathbb{N}_0$ denotes the finite planning horizon.

For the above POMDP, at each time step, the environment is in some state $s \in \states$ and the agent interacts with the environment by taking an action $a \in \actions$. Doing so results in the environment transitioning to a new state $\bar{s} \in \states$ in the next time step with probability $T(s,a,\bar{s})$. In a POMDP the agent doesn't have access to the true state of the environment. Instead, it receives an observation $o \in \obs$ regarding the state of the environment with probability $Z(o|\bar{s},a)$ which depends on the new state of the environment and the action taken. The agent can update its belief about the true state of the environment using this observation. The agent also receives a reward $R(s,a)$. 

The problem of optimal policy synthesis for a finite-horizon POMDP is that of choosing a sequence of actions which maximizes the expected total reward $\EE[\sum_{t=0}^{H}r_t]$ where $r_t$ is the reward earned at time instant $t$.  
Hence the optimal behavior may often include actions which are taken simply because they improve the agent's belief about the true state, e.g., an inspection action taken to learn the condition of a component.

\subsection{Data-Driven Approaches for Decision Making}

Data-driven methodologies, particularly reinforcement learning (RL) and imitation learning (IL), have emerged as powerful tools for sequential decision-making under uncertainty \cite{sutton2018reinforcement}, \cite{hussein2017imitation}. In the context of infrastructure management, these approaches offer the potential to develop adaptive policies that can handle the stochastic deterioration of components and partial observability of their conditions.

In RL, an agent learns to make optimal decisions by interacting with an environment to learn a policy $\pi: \states \rightarrow \actions$ that maximizes the expected cumulative reward over a horizon $H$:

\begin{equation} \pi^* = \argmax_{\pi} \EE_{\pi}\left[\sum_{t=0}^{H} R(s_t, \pi(s_t))\right] \end{equation} where $s_t \in \states$ is the state at time $t$, and $R(s_t, a_t)$ is the reward obtained by taking action $a_t$ in state $s_t$. RL algorithms, such as Q-learning \cite{watkins1992q} and policy gradient methods \cite{williams1992simple}, enable the agent to learn optimal policies without requiring explicit models of the environment's dynamics.

However, applying RL to infrastructure management poses challenges due to the complexity and scale of real-world systems. The large size of the state and action spaces, along with the partial observability of component conditions, complicate the learning process. Moreover, collecting sufficient real-world data for training is often impractical or costly.

Imitation learning addresses some of these challenges by leveraging expert demonstrations to learn effective policies. Given a set of expert trajectories $\mathcal{D} = \{(s_0, a_0), (s_1, a_1), \ldots, (s_T, a_T)\}$, IL methods aim to learn a policy that mimics the expert's behavior. Behavioral cloning treats this as a supervised learning problem, minimizing the discrepancy between the agent's actions and the expert's actions:
\begin{equation} \pi^* = \argmin_{\pi} \sum_{(s,a) \in \mathcal{D}} \ell(\pi(s), a) \end{equation} where $\ell$ is a loss function measuring the difference between the predicted action $\pi(s)$ and the expert action $a$. This approach can be advantageous in infrastructure management, where expert knowledge and historical data are available.

Inverse reinforcement learning (IRL) goes a step further by seeking to infer the underlying reward function that the expert is optimizing \cite{ng2000algorithms}. By recovering the reward function $R(s, a)$ from expert demonstrations, IRL enables the development of policies that generalize beyond the observed behavior and can adapt to new situations.

The utility of RL, IL, and IRL in infrastructure management depends heavily on the availability of accurate simulation environments and high-quality expert data. Developing realistic simulation tools that capture the stochastic dynamics and partial observability inherent in infrastructure systems is essential. Furthermore, standard benchmarks and baselines are needed to evaluate and compare different approaches consistently.

To harness data-driven methods for infrastructure management, it is essential to formalize the problem in a way that captures its inherent complexities, such as stochastic deterioration and partial observability. In the next section, we model the infrastructure management problem as a Partially Observable Markov Decision Process and outline the research challenges that this formulation presents.

\section{POMDP Model of Infrastructure}
\label{sec:problem_statement}
In this section, we focus on modeling infrastructure management problems that involve stochastic deterioration and partial observability, requiring decisions to be made under budget constraints. We frame these problems using Partially Observable Markov Decision Processes (POMDPs), which provide a powerful framework for modeling decision-making under uncertainty and incomplete information. We then discuss the research problems that can be studied within this paradigm.

We consider a large-scale infrastructure system comprising $n$ components, each subject to deterioration over time. We assume knowledge about each component's deterioration dynamics, possible maintenance actions, associated costs, and the overall budget constraints. Our goal is to model the decision-making process for managing these components effectively under uncertainty and partial observability.

We model the infrastructure system as a collection of independent POMDPs, one for each component, with a shared budget constraint that couples their decision processes. Each component's POMDP captures its individual deterioration dynamics and the effects of maintenance actions. In a later section, we will discuss how we implement this modeling approach in our tool, \textit{InfraLib}.

We now present the general formulation of modeling a component as a POMDP, detailing its key elements.

\paragraph{States:} The state space $\mathcal{S}$ of each component represents its condition, typically quantified by a condition index (CI). The CI can take discrete values from $0$ (worst condition) to $s_{max}$ (best condition). The true state of a component is typically not directly observable without inspection.

\paragraph{Observations:} The observation space $\mathcal{O}$ includes possible observed conditions of the component after certain actions and a null observation when no information is obtained. Observations provide potentially noisy information about the component's true state.

\paragraph{Actions:} The action space $\mathcal{A}$ includes any possible maintenance decisions that can be taken in the infrastructure system, including (i) \texttt{Do Nothing}, an action that lets the component deteriorate naturally without any intervention, (ii) \texttt{Inspect}, an action that allows the agent to obtain the true state of the component without affecting its condition, (iii) \texttt{Repair}, an action that improves the component's condition by a certain amount, and (iv) \texttt{Replace}, an action that restores the component to its best possible condition. Each action has associated costs and effects on the component's state and observations.

\paragraph{Transitions:} The transition model defines the probabilities of moving between states given an action. Deterioration is stochastic, and the transition probabilities capture the uncertainty in how the component's condition changes over time. For example, selecting the \texttt{Do Nothing} action may result in the component's condition worsening according to a probabilistic deterioration model.

\paragraph{Rewards:} The reward function reflects the objectives of the management problem, such as maximizing the infrastructure's overall condition, minimizing failures, or optimizing cost-effectiveness. The specific reward structure depends on the particular problem being addressed. In our framework, rewards can be defined to capture various objectives, and we provide examples in a later section. Additionally, rewards can be aggregated across components and time to model the overall performance of the infrastructure system under budget constraints.

In the following section, we discuss the research problems that can be modeled and studied within this framework.
\section{Planning Problems for Infrastructure Management}
\label{sec:research_problems}

Modeling infrastructure management using POMDPs opens up several research avenues that address key challenges in the field. Some of the important research problem studied within this paradigm include:

\subsubsection*{Optimal Maintenance Scheduling under Budget Constraints}

Determining the optimal allocation of limited resources for maintenance, repair, and replacement actions across multiple components is a critical problem. The goal is to maximize the overall infrastructure performance or minimize the risk of failures while adhering to budget constraints. Research focuses on developing policies that balance short-term costs with long-term benefits, taking into account the stochastic nature of deterioration and partial observability.

\subsubsection*{Planning under Partial Observability}

Since the true states of components are not always known without inspection, decision-making must account for uncertainty in component conditions. Developing algorithms that effectively handle partial observability and make informed decisions based on beliefs about the component states is essential. Research explores advanced POMDP solution methods, belief state representations, and efficient planning algorithms tailored to infrastructure management.

\subsubsection*{Stochastic Deterioration Modeling}

Accurately modeling the stochastic deterioration of components is crucial for reliable decision-making. Research focuses on improving the transition models to better reflect real-world deterioration processes, incorporating factors such as environmental conditions, usage patterns, and historical data. Approaches include learning or estimating transition probabilities from data and updating models as new information becomes available.

\subsubsection*{Scalability and Computational Efficiency}

Infrastructure systems often involve a large number of components, making the computational complexity of solving the POMDPs a significant challenge. Research addresses scalability by developing approximate solution methods, decomposition techniques, or hierarchical approaches that reduce computational demands while maintaining acceptable performance.

\subsubsection*{Multi-Agent Coordination and Hierarchical Decision-Making}

In practice, decisions in infrastructure management may be made at different hierarchical levels, with coordination among multiple decision-making agents. Research explores multi-agent POMDP frameworks, hierarchical planning, and coordination mechanisms that enable effective decision-making across different levels of the infrastructure system.

\subsubsection*{Robustness to Model Change and Uncertainty}

Given the potential discrepancies between modeled deterioration dynamics and real-world behavior, it is important to develop policies that are robust to model changes and uncertainty. Research investigates methods for policy robustness, adaptive learning, and techniques for transferring policies from simulation to real-world deployment (sim-to-real transfer), ensuring that solutions remain effective in practice.

\subsubsection*{Incorporating Risk and Reliability Measures}

Infrastructure management often requires consideration of risk and reliability metrics, such as the probability of component failure or system-level performance thresholds. Research focuses on integrating risk assessments into the POMDP framework, enabling decision-making that accounts for both expected rewards and risk mitigation.

By providing a comprehensive toolkit that implements this POMDP framework and scales effectively to real-world infrastructure systems, we can enable researchers from a wide range of disciplines to focus on specific components in the abstraction while analyzing their broader impacts across diverse applications. To facilitate this research, we introduce InfraLib --- our implementation that brings together these modeling capabilities and a suite of solution approaches.

\section{InfraLib}
\label{sec:infralib}
InfraLib is a comprehensive modeling, simulation, and analysis framework designed to enable research into data-driven decision making for infrastructure management under uncertainty. It provides predefined, structured environments while also allowing users to flexibly define custom scenarios and constraints. The code, documentation, example environments, and tutorials are available as an open-source library under the permissive MIT license at \url{infralib.github.io}.

\subsection{InfraLib Structure}
InfraLib framework adopts a modular architecture, which enables separation of concerns and easy extensibility. The core infrastructure model is designed to be highly customizable, allowing users to define custom components, deterioration models, objectives, constraints, and management actions. The hierarchical structure of infrastructure systems is also configurable, enabling users to group components into units and facilities in domain-specific ways.

At the core of InfraLib is the infrastructure simulator that requires three key modules:
\begin{itemize}
    \item \texttt{DynamicsModel}, which defines the stochastic deterioration dynamics of each component, 
    \item \texttt{CostModel}, which defines the costs of each action, and 
    \item \texttt{BudgetModel}, which defines the budget constraints and allocation mechanisms.
\end{itemize}
In addition to these modules, InfraLib also provides several abstractions that enable modeling nuances of real-world systems. These modules include:
\begin{itemize}
    \item \texttt{HierarchyModule}, which defines the hierarchical structure of the infrastructure system and relevant hierarchical actions, and
    \item \texttt{MetadataModule}, which provides a flexible way to attach actionable metadata to components, actions, and other entities in the infrastructure system.
\end{itemize}
These modules are combined to generate a POMDP environment that powers the simulation, user interface, analysis, and RL environments. InfraLib makes minimal assumptions about the specific structure of these models, while also providing strong defaults for each model. This design allows users focused on say devloping algorithms to easily integrate and test their approaches, while allowing users with expertise in infrastructure management to customize the models to their needs. 

\subsection{InfraLib Default Models}
The default models in InfraLib are carefully chosen based on established infrastructure management literature. For the dynamics model, we implement a Weibull-based deterioration model following \cite{grussing2006condition}, where the Condition Index (CI) of each component evolves according to a Markov chain with transition probabilities derived from Weibull distributions. For component $i$, the CI at time $t$ is given by
\begin{equation}
    \text{CI}^i(t) = \left\lfloor 100 \times \left(1 - \left(1 - e^{-\left(\frac{t}{\lambda^i}\right)^{k^i}}\right)\right)\right\rfloor,
\end{equation}
where normally distributed $k^i$ and $\lambda^i$ are the shape and scale parameters respectively. This formulation captures the natural deterioration pattern where components start at CI = 100 (perfect condition) and gradually degrade towards CI = 0 (complete failure). The stochastic nature of the parameters accounts for variations in deterioration patterns observed in real infrastructure systems.

The default cost model implements a nonlinear cost function that reflects the empirically observed relationship between component condition and repair costs \cite{zhao2019optimal}. For component $i$, the repair cost $c_r^i$ is given by
\begin{equation}
    c_r^i = \left(\frac{100 - s^i}{100 - \delta^i}\right)^{\alpha^i} \times c_m^i + \beta^i c_m^i,
\end{equation}
where $s^i$ is the current CI, $\delta^i$ is the failure threshold, $\alpha^i$ is a cost-sensitivity parameter, $c_m^i$ is the replacement cost, and $\beta^i$ is the minimum repair cost fraction. This formulation captures the practical reality that repairs become disproportionately expensive as components approach failure, while also accounting for the minimum cost involved in any repair action.

For budget constraints, InfraLib provides two default models: a fixed budget model for scenarios with predetermined initial budget, and a cyclic budget model which captures periodic budget allocations common in public infrastructure management \cite{huang2010budget}. The cyclic budget model supports variable cycle lengths and amounts, defined as
\begin{equation}
    B(t) = b_k \text{ where } k = \arg\max_j \left\{t \geq t_j\right\},
\end{equation}
where $b_k$ is the budget for cycle $k$ and $t_j$ are cycle start times. This flexibility allows modeling of both fiscal year allocations and more complex multi-year funding patterns observed in practice.

These default models provide a strong foundation to jumpstart research in decision-making at infrastructure scalewhile also serving as templates for users to build their own models. InfraLib documentation provides detailed guides on how to customize these models to specific scenarios and applications.

InfraLib is implemented as a Python library, leveraging NumPy and Numba for efficient vectorized computation. The framework is designed to be user-friendly, with a simple and intuitive API that abstracts the underlying complexity. InfraLib also provides a suite of tools for analysis, visualization, storage, and retrieval of data. A key emphasis in InfraLib's design is scalability and computational efficiency. The framework can simulate infrastructure systems comprising millions of components and spanning long time horizons. This massive scale is crucial for bridging the gap between research and the complexity of real-world infrastructure networks.

\subsection*{RL Environments}

InfraLib provides RL and decision-making environments by wrapping its core dynamics simulation with a Gymnasium-like interface \cite{towers2024gymnasium}. This design allows users to interact with the simulation using a standardized interface with methods such as \texttt{reset()}, \texttt{step(action)}, and \texttt{render()}, facilitating seamless integration with popular RL libraries.

The reward function $\mathcal{R}$ is customizable and can be defined based on the system state, action, or both, aligning with the specific objectives of the problem. This flexibility enables optimization for various goals, such as minimizing maintenance costs or maximizing infrastructure reliability.

The observation space provided to the agent in the environment is also configurable. It can represent the full simulation state, partial observations, or any function of the state or observations, depending on user requirements. Such flexibility accommodates scenarios with full observability, partial observability, or state abstractions, reflecting a wide range of real-world situations.

InfraLib offers environment templates that are directly compatible with Stable-Baselines3 \cite{stable-baselines3} --- a popular library covering all popular RL algorithms --- enabling easy application and comparison of different RL solution methods. Each environment includes a unified evaluation interface and problem-specific metrics, which assist in comparing the performance of learning and non-learning methods on consistent benchmarks.

\begin{figure*}[t]
    \centering
    \begin{subfigure}[b]{0.48\textwidth}
        \centering
        \includegraphics[width=\textwidth]{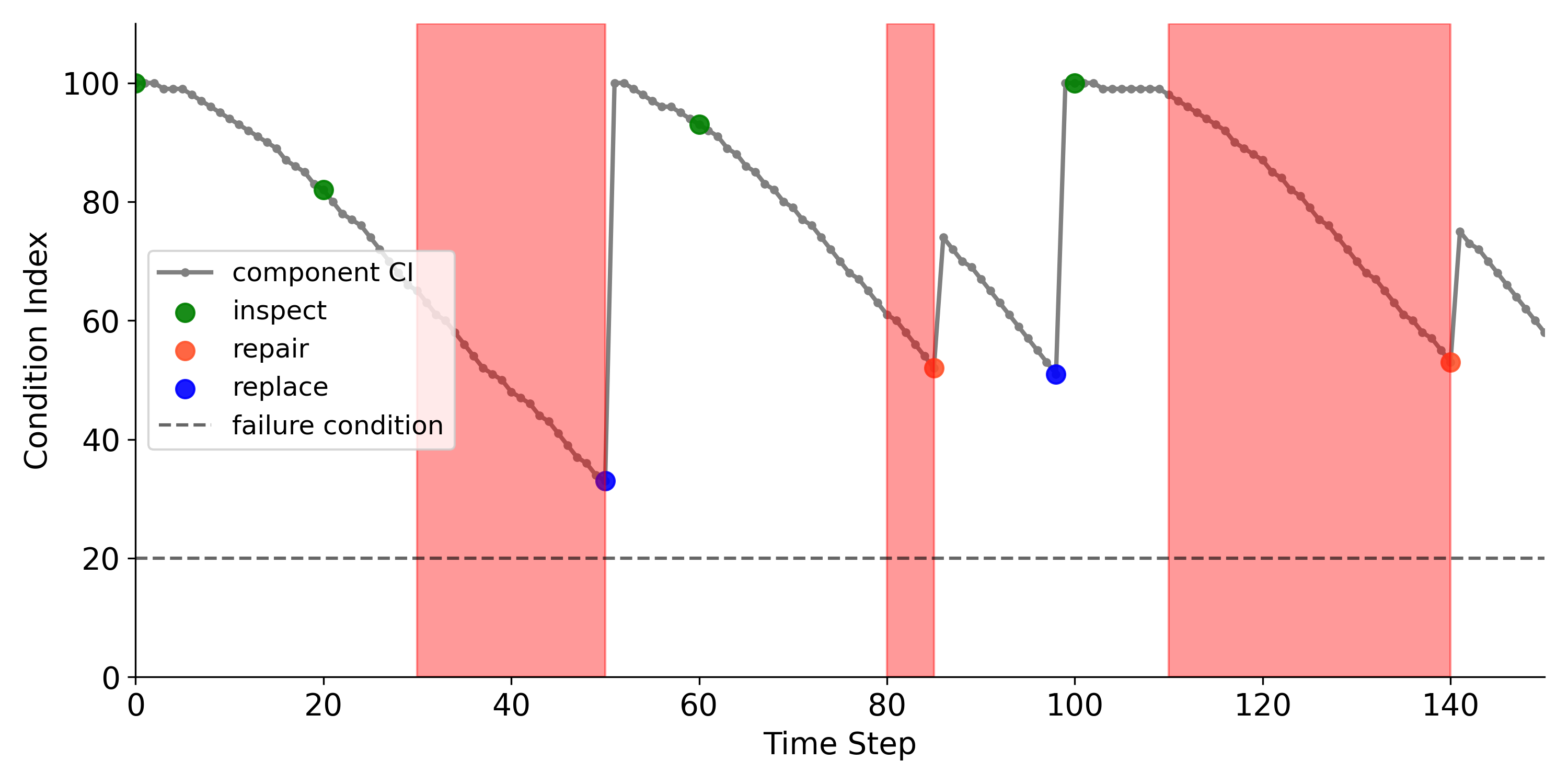}
        \caption{}
        \label{fig:intermittent_availability}
    \end{subfigure}
    \hfill
    \begin{subfigure}[b]{0.48\textwidth}
        \centering
        \includegraphics[width=\textwidth]{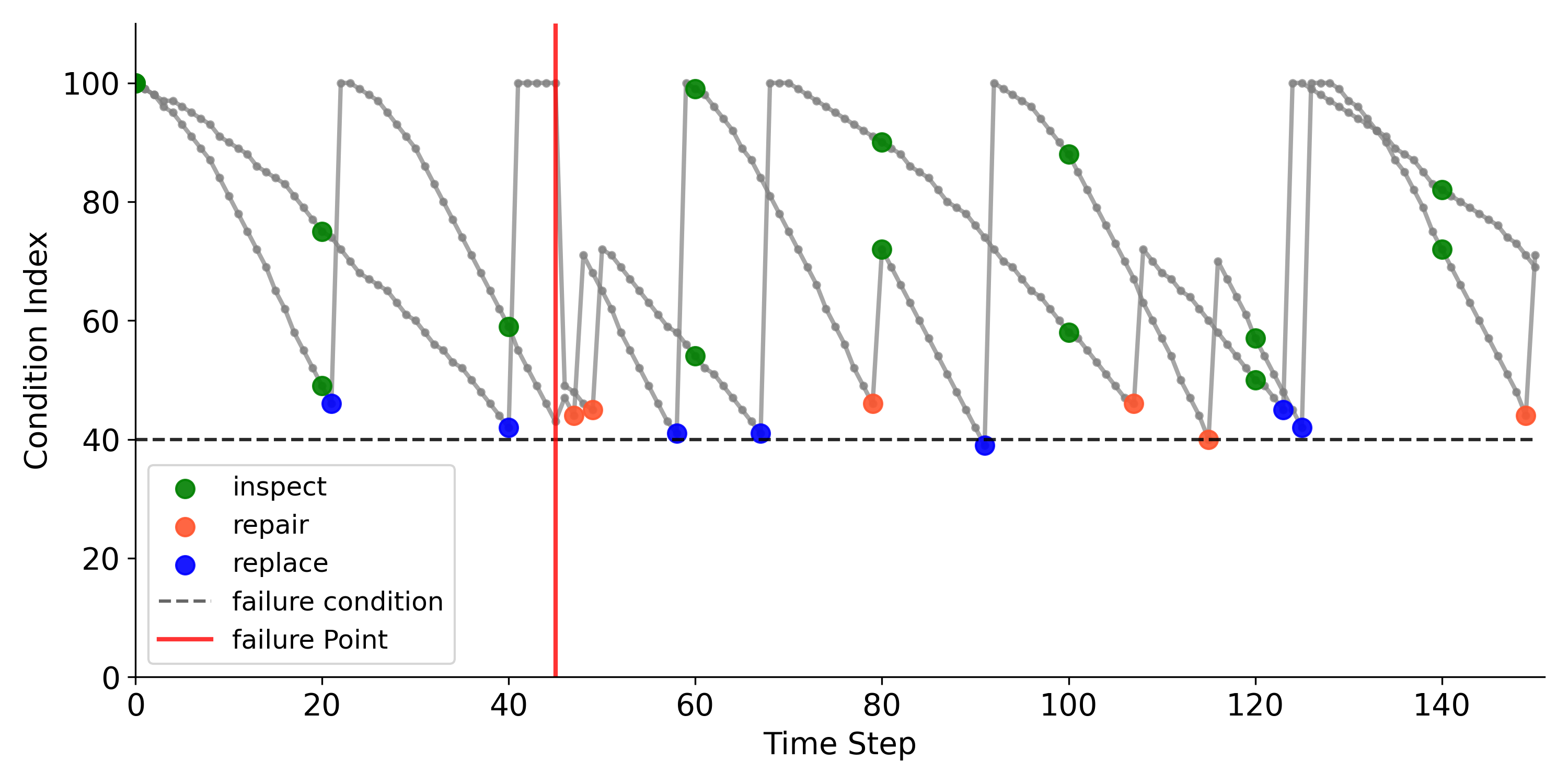}
        \caption{}
        \label{fig:catastrophic_failure}
    \end{subfigure}
    \caption{Component condition index over time from InfraLib simulations illustrating (a) intermittent unavailability periods (red regions) and (b) catastrophic failure (at time step 45).}
    \label{fig:combined}
\end{figure*}

We provide a series of predefined environments covering a variety of scenarios:

\begin{itemize}
    \item \textbf{Simple Infrastructure Management Environment}: A basic setup with a limited number of components and straightforward deterioration dynamics, ideal for testing algorithms and educational purposes.
    \item \textbf{Large-Scale Infrastructure Environment}: Simulates extensive networks with hundreds of thousands of components to test scalability and computational efficiency.
    \item \textbf{Cyclic Budget Environment}: Incorporates periodic budget constraints to reflect common financial cycles in infrastructure management.
    \item \textbf{Catastrophic Failure Environment}: Introduces sudden failure events affecting multiple components, useful for stress-testing policies under extreme conditions and modeling extreme weather events scenarios, as illustrated in Figure \ref{fig:catastrophic_failure}.
    \item \textbf{Intermittent Component Availability Environment}: Accounts for components that are only available for maintenance during specific time windows, as illustrated in Figure \ref{fig:intermittent_availability}.
\end{itemize}

These environments are built using InfraLib's features—such as flexible dynamics, cost models, and budget constraints—and serve as testbeds for RL algorithms in infrastructure management.

In addition to algorithmic approaches, human expertise is often essential for validating and refining infrastructure management strategies. In the next section, we discuss how InfraLib supports human-in-the-loop decision-making, data collection from experts, and scenario exploration.

\subsection{InfraLib Human Interface}
\label{sec:human_interface}

In addition to providing decision-making environments, InfraLib functions as a powerful analysis tool and a platform for capturing human expert actions to understand decision-making in complex infrastructure management scenarios. The simulator facilitates exploration of various management strategies, enabling researchers and practitioners to analyze and compare human decisions under different conditions.

\begin{figure}[h]
    \centering
    \includegraphics[width=0.9\textwidth]{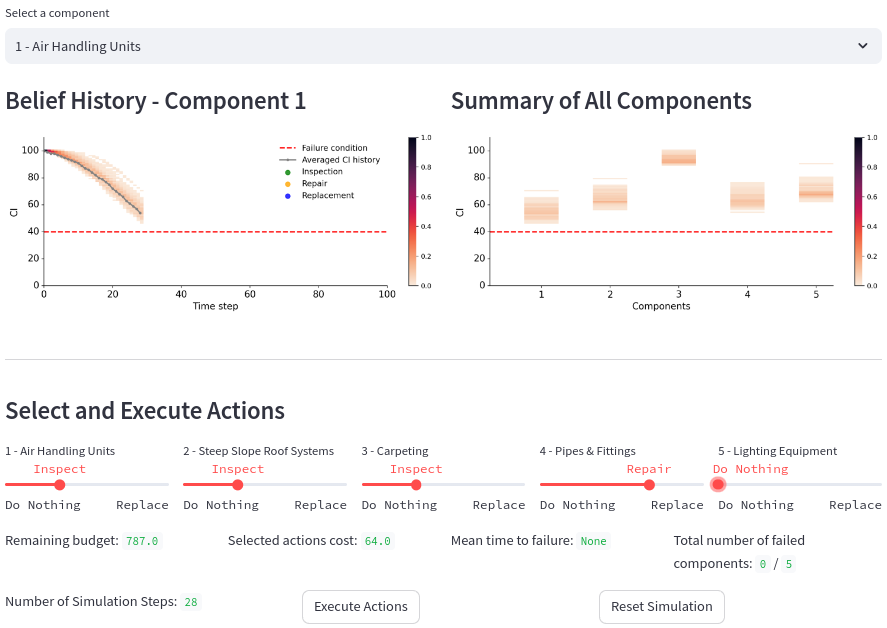}
    \caption{Snapshot from InfraLib's analysis and data collection human interface.}
    \label{fig:dashboard}
\end{figure}

InfraLib's human interface is powered by an intuitive web-based graphical user interface (GUI) that allows experts to interact with simulated infrastructure systems. Through this interface, experts can test different management strategies and quickly analyze "what-if" scenarios. The GUI provides corresponding risk analyses by evaluating simulations over a range of parameters, offering insights into the potential outcomes of various decisions. Figure \ref{fig:dashboard} shows a snapshot of the GUI interface. A web-demo of the human interface is available at \url{infralib.github.io}.

In addition to analyzing the performance of different strategies, users can also employ the human interface to create a variety of infrastructure management scenarios and capture the actions of human experts, leveraging their domain knowledge. The interface logs all relevant information seamlessly, including expert actions, observations, and environment states. This data can potentially be utilized with imitation learning algorithms to derive policies that mimic expert behavior, or with inverse reinforcement learning to infer the underlying reward functions guiding expert decisions. This approach enables the development of models that closely align with human expertise without requiring explicit reward definitions or extensive exploration.

InfraLib also supports the injection of human knowledge and overrides into simulations. Experts can specify custom policies, thresholds, or strategies for managing specific components, allowing for human-in-the-loop systems where experts guide and correct a deployed data-driven policy. This bidirectional interaction has the potential to enhance both human and automated decision-making, fostering collaborative approaches to infrastructure management.

To demonstrate the practical applications and versatility of InfraLib, we now present several example environments, benchmarks, and case studies that cover a variety of scenarios discussed in earlier sections.

\section{Experiments and Case Studies}
\label{sec:example_environments}
This section presents sample environments that demonstrate InfraLib's modeling capabilities. The section has two primary objectives: (i) to establish a set of ready-to-use RL environments for decision-making research and benchmarking, (ii) to showcase case studies that illustrate InfraLib's versatility and serve as templates for custom environment development. We examine three distinct scenarios: a reinforcement learning policy for managing five infrastructure components, a large-scale benchmark system with 100,000 components, and a real-world case study of pavement maintenance in Manhattan's road network.

\subsection*{RL Policy for Infrastructure Management}
We consider a simple environment with 5 components and a fixed budget of 2000 units. Each component has a failure threshold of 40 CI, below which the component is considered to have failed \cite{grussing2006condition}. The environment uses InfraLib's default Weibull dynamics model for component deterioration, default cost model for action costs, and a fixed budget model for budget management.

For each component, the agent can take one of four actions: \texttt{Do Nothing}, \texttt{Inspect}, \texttt{Repair}, or \texttt{Replace}. The objective is to maximize the number of time steps before any component fails, with a maximum episode length of 100 steps. This represents a simplified yet challenging infrastructure management scenario where the agent must balance the use of limited resources across multiple components while preventing failures under partial observability.

We model this as a RL environment where the agent receives an observation vector $o_t = [o^1_t, \ldots, o^n_t, b_t, \tau^1_t, \ldots, \tau^n_t]$ at each time step $t$, where $o^i_t \in [0,101]$ represents the last observation for component $i$ (with 101 indicating no observation), $b_t$ is the remaining budget, and $\tau^i_t \in [0,100]$ is the time since the last inspection for component $i$. The action space $\mathcal{A}$ is discrete with size equating $4^5$, representing all possible combinations of the four actions across $5$ components.

The illustrative reward function is designed to encourage maintaining components well above their failure thresholds:
\begin{equation}
r_t = \frac{1}{100n}\sum_{i=1}^n (s^i_t - \delta^i) - 10\mathbbm{1}_{\{s^i_t \leq \delta^i\}} 
\end{equation}
where $s^i_t$ is the CI of component $i$ at time $t$, $\delta^i = 40$ is the failure threshold, and $\mathbbm{1}$ is the indicator function.

We evaluate three reinforcement learning algorithms --- PPO, A2C, and, DQN --- using the Stable-Baselines3 interface. Each algorithm is trained for 2 million time steps with three different random seeds. Detailed hyperparameters for each algorithm are available in the InfraLib library. For comparison, we implement two baseline policies:
\begin{enumerate}
    \item No-Action baseline: Takes no actions, allowing natural deterioration of the components
    \item Rule-Based baseline: Inspects each component every 5 time steps and performs repairs if a component's CI is within 20 units of its failure threshold after inspection
\end{enumerate}

\begin{figure}[h]
    \centering
    \includegraphics[width=\textwidth]{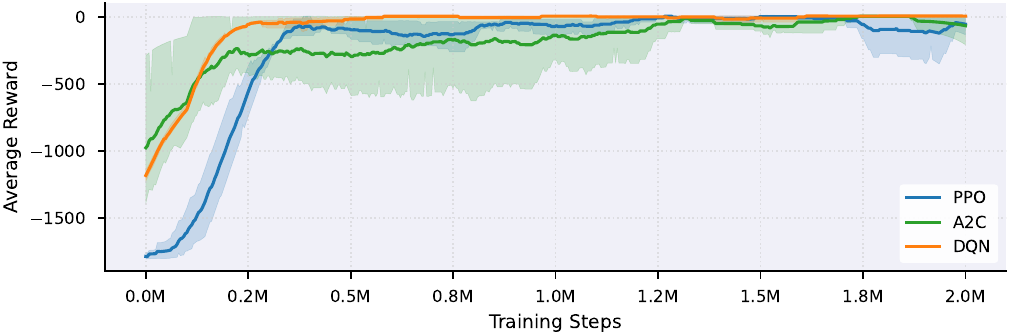}
    \caption{Average reward values during training for the three evaluated RL algorithms.}
    \label{fig:rl_training}
\end{figure}

\begin{figure}[h]
    \centering
    \includegraphics[width=\textwidth]{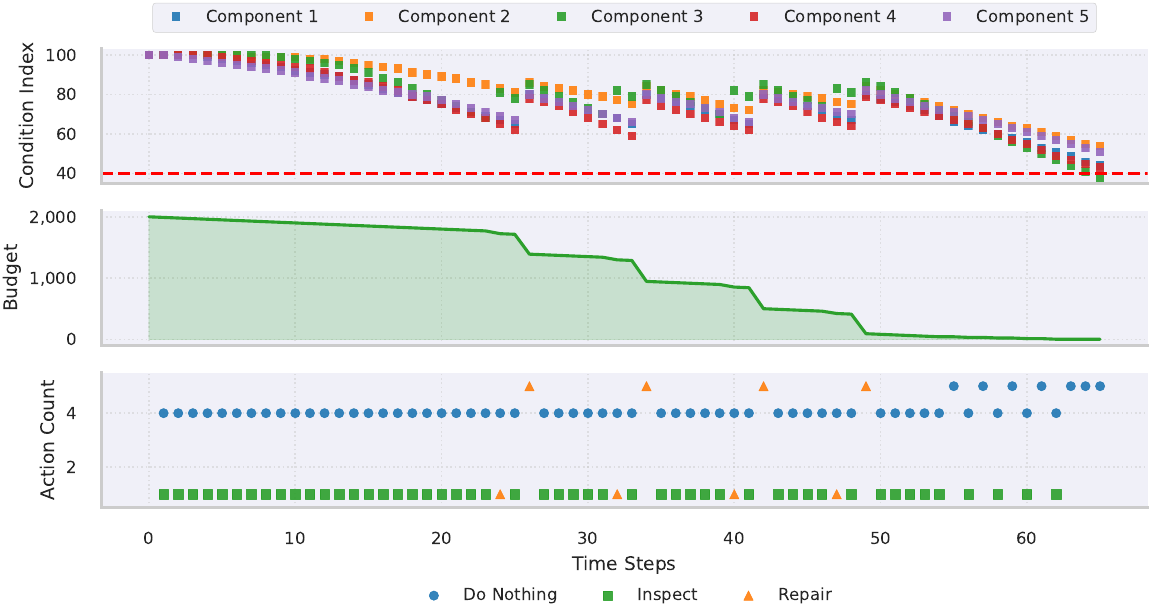}
    \caption{A sample rollout of the DQN policy on the environment with five components.}
    \label{fig:rl_evaluation}
\end{figure}

Figure \ref{fig:rl_training} shows the training curves for all three RL algorithms averaged across the three seeds. DQN demonstrates the fastest convergence and most stable learning, with PPO following closely behind. Figure \ref{fig:rl_evaluation} illustrates a representative rollout of the DQN policy. The policy efficiently manages the budget taking cheaper but informative inspection actions in the early stages, before performing repairs when necessary. The DQN policy achieves a 54\% longer average time to first component failure compared to the rule-based baseline and a 103\% longer average time to failure compared to the no-action baseline.

\subsection{LargeSys-100K - Large-Scale Infrastructure Management Benchmark}
Real-world infrastructure systems often comprise hundreds of thousands of components, making scalability a critical requirement for practical management solutions. To facilitate research on such large-scale systems and demonstrate InfraLib's scalability, we introduce LargeSys-100K, a benchmark environment containing 100,000 component instances across 100 different component types, with 1,000 instances per type.

LargeSys-100K employs the same default models as the previous environment: Weibull dynamics model for deterioration, default standard cost model for action costs, and a fixed budget for budget management. The parameters for these models are carefully chosen based on empirical ranges observed in real-world infrastructure component data \cite{grussing2006condition,vora2023welfare}.

We consider the problem of managing this system with a budget of 20,000,000 where components fail at CI = 0. The primary performance metric is the Expected Time to Failure (ETTF), which measures the average time to failure across all components.

This benchmark presents significant computational challenges that make it unsuitable for direct application of standard RL approaches. The action space dimension is $4^{100000}$, making it infeasible to solve as a single POMDP as in the previous section. Moreover, the large state space and number of components pose substantial challenges for memory usage and computation time, particularly when maintaining belief states and computing value functions. These challenges make LargeSys-100K an ideal testbed for developing and evaluating scalable approaches to infrastructure management.

To establish baseline performance on this benchmark, we provide several reference implementations. The first is a no-action policy that represents the default system behavior without any maintenance interventions. We also implement four variants of a rule-based policy that inspects components periodically and takes actions based on predefined CI thresholds. These variants are parameterized by two key factors: the inspection interval $\tau$ and the failure threshold $\theta$. The inspection interval $\tau$ determines how frequently components are checked, while $\theta$ defines the CI threshold below which components are replaced. Finally, we implement the approach proposed in \cite{vora2024solving}, which we refer to as Oracle-Guided Meta-RL (OGM-RL). This method first uses a random forest model to distribute the shared budget among components by approximating their optimal value functions, followed by an oracle-guided meta-trained PPO algorithm that solves the resulting single-component POMDPs independently.

\begin{table}[h]
    \centering
    \caption{Performance Comparison on LargeSys-100K Benchmark}
    \label{tab:largesys_results}
    \begin{tabular}{lccc}
        \toprule
        Method & ETTF & Budget Utilized (\%) & Replacements (total) \\
        \midrule
        No-Action & 14.8 & 0.0 & 0 \\
        RB-1-10 & 15.9 & 100.0 & 6259 \\
        RB-5-10 & 19.7 & 100.0 & 39569 \\
        RB-10-10 & 16.2 & 39.6 & 18737 \\
        RB-10-20 & 22.5 & 92.8 & 57977 \\
        OGM-RL & \textbf{34.3} & 36.3 & 33527 \\
        \bottomrule
    \end{tabular}
    \caption*{\small Note: Results averaged over 100 independent runs. ETTF: expected time to failure. RB-$\tau$-$\theta$ denotes rule-based policy with inspection interval $\tau$ and failure threshold $\theta$.}
\end{table}

Table \ref{tab:largesys_results} compares the performance of the provided baselines. The rule-based approaches show moderate improvements over the no-action baseline, with RB-10-20 achieving the best performance among the variants. OGM-RL significantly outperforms all rule-based approaches, achieving the highest ETTF value of 34.3. However, the relatively low budget utilization suggests that there is still substantial room for improvement. 

\subsection*{Manhattan Road Network Maintenance}
Finally, we demonstrate InfraLib's capabilities to model real-world infrastructure management problems using a pavement maintenance task in a section of Manhattan, New York City, an urban environment characterized by high traffic density. The study area is based on \cite{blahoudek2022efficient}, and is bounded by 2nd Avenue (east), 11th Avenue/West End Avenue (west), 42nd Street (south), and 116th Street (north), represented as a network with 1024 vertices and 2118 edges sourced from OpenStreetMap data.

Using InfraLib, we model each road segment as a component by a Weibull deterioration dynamics model, where the parameters of the dynamics are determined based on the segment's average speed data \cite{blahoudek2021fuel,zheng2021optimal}. This modeling choice is motivated by urban traffic phenomena: in Manhattan's dense environment, lower average speeds typically indicate higher traffic volumes and more frequent stop-and-go patterns, leading to accelerated pavement wear through increased vehicle interactions and repetitive stress from braking and acceleration forces \cite{yazici2017urban}. We note that the generated deterioration dynamics are illustrative and are not intended to be an accurate representation or predictive model of real-world system behavior.
\begin{figure}
    \centering
    \includegraphics[width=\textwidth]{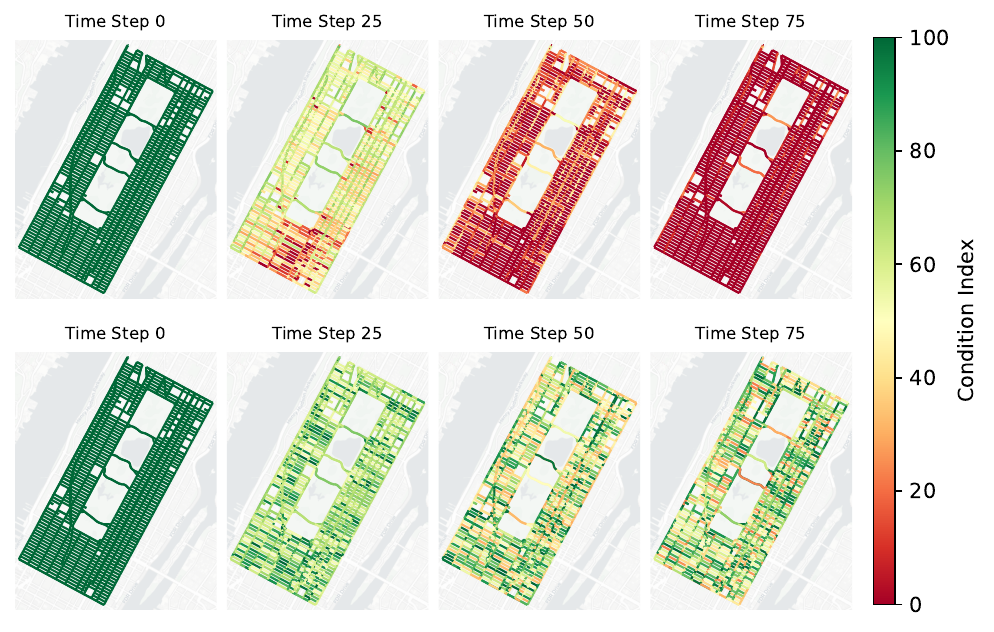}
    \caption{Simulated condition indices of Manhattan's road network using InfraLib. The visualization shows network conditions at simulation steps 0, 25, 50, and 75, comparing scenarios without maintenance (top) and with maintenance under a fixed cyclic budget allocation (bottom).}
    \label{fig:manhattan_deterioration}
\end{figure}
The maintenance model follows a cyclic budget, where a fixed amount of budget gets allocated every budget cycle. The maintenance actions available for each road segment include minor repairs (improving condition by 30\%) and major rehabilitation (restoring to perfect condition) with costs proportional to the segment length with a constraint on the total amount of road segments that can be rehabilitated in each time step.

Figure \ref{fig:manhattan_deterioration} visualizes the simulation results over a 75-step period under two scenarios: no maintenance and with a greedy maintenance policy that performs maintenance on road segments based on their condition index and importance stores. The deterioration patterns reveal the impact of no maintenance, with high-traffic segments showing accelerated deterioration. The maintenance scenario demonstrates InfraLib's ability to integrate real-world infrastructure data with customized deterioration models and maintenance constraints, providing a flexible framework for studying infrastructure management strategies in urban environments.

\section{Conclusion and Future Work}
This paper introduced InfraLib, a comprehensive framework for modeling and analyzing large-scale infrastructure management problems under uncertainty. The framework makes four key contributions: a flexible, hierarchical modeling approach that captures the stochastic nature of infrastructure deterioration and partial observability; scalable simulation capabilities that enable modeling of real-world systems with hundreds of thousands of components; a graphical user interface for human-in-the-loop decision-making; and standardized environments and benchmarks for evaluating learning-based approaches. Through detailed case studies on both synthetic benchmarks and real-world systems, we demonstrated InfraLib's ability to model diverse scenarios ranging from simple maintenance scheduling to complex, large-scale asset management problems.

Several promising directions exist for future research and development of InfraLib. A key extension would be incorporating explicit modeling of maintenance crew availability and scheduling to better reflect real-world operational constraints, including resource mobilization costs and emergency repairs. Additionally, the framework could be enhanced to support other learning approaches such as meta-learning for rapid adaptation to new infrastructure types and transfer learning across different domains. A tighter integration with geographic information systems and existing asset management software would further bridge the gap between research and practice.

\section{Data Availability Statement}
All code, models, and documentation for InfraLib described in this study are openly available in a GitHub repository at infralib.github.io under the MIT license. The Manhattan road network data used in the case study was obtained from OpenStreetMap, which is freely available at openstreetmap.org. The synthetic benchmark environments, including LargeSys-100K, and all associated configuration files are included in the InfraLib repository.

\bibliography{references}

\begin{thebibliography}{36}
\providecommand{\natexlab}[1]{#1}
\providecommand{\url}[1]{\texttt{#1}}
\expandafter\ifx\csname urlstyle\endcsname\relax
  \providecommand{\doi}[1]{doi: #1}\else
  \providecommand{\doi}{doi: \begingroup \urlstyle{rm}\Url}\fi

\bibitem[Egert et~al.(2009)Egert, Kozluk, and Sutherland]{egert2009infrastructure}
Balazs Egert, Tomasz~J. Kozluk, and Douglas Sutherland.
\newblock Infrastructure and growth: empirical evidence.
\newblock Working Paper 2700, CESifo Working Paper, July 2009.

\bibitem[Srinivasu and Rao(2013)]{srinivasu2013infrastructure}
Bathula Srinivasu and Srinivasa Rao.
\newblock Infrastructure development and economic growth: Prospects and perspective.
\newblock \emph{Journal of Business Management and Social Sciences Research}, 2\penalty0 (1):\penalty0 81--91, 2013.

\bibitem[Yang et~al.(2018)Yang, Ng, Xu, and Skitmore]{yang2018towards}
Yifan Yang, Thomas Ng, Frank Xu, and Martin Skitmore.
\newblock Towards sustainable and resilient high density cities through better integration of infrastructure networks.
\newblock \emph{Sustainable Cities and Society}, 42:\penalty0 407--422, 2018.

\bibitem[Zimmerman(2001)]{zimmerman2001social}
Rae Zimmerman.
\newblock Social implications of infrastructure network interactions.
\newblock \emph{Journal of Urban Technology}, 8\penalty0 (3):\penalty0 97--119, 2001.

\bibitem[Rinaldi et~al.(2001)Rinaldi, Peerenboom, and Kelly]{rinaldi2001identifying}
Steven~M Rinaldi, James~P Peerenboom, and Terrence~K Kelly.
\newblock Identifying, understanding, and analyzing critical infrastructure interdependencies.
\newblock \emph{IEEE Control Systems Magazine}, 21\penalty0 (6):\penalty0 11--25, 2001.

\bibitem[Kabir et~al.(2014)Kabir, Sadiq, and Tesfamariam]{kabir2014review}
Golam Kabir, Rehan Sadiq, and Solomon Tesfamariam.
\newblock A review of multi-criteria decision-making methods for infrastructure management.
\newblock \emph{Structure and Infrastructure Engineering}, 10\penalty0 (9):\penalty0 1176--1210, 2014.

\bibitem[Madanat(1993)]{madanat1993optimal}
Samer Madanat.
\newblock Optimal infrastructure management decisions under uncertainty.
\newblock \emph{Transportation Research Part C: Emerging Technologies}, 1\penalty0 (1):\penalty0 77--88, 1993.

\bibitem[Frangopol and Bocchini(2019)]{frangopol2019bridge}
Dan Frangopol and Paolo Bocchini.
\newblock Bridge network performance, maintenance and optimisation under uncertainty: Accomplishments and challenges.
\newblock \emph{Structures and Infrastructure Systems}, pages 30--45, 2019.

\bibitem[Makovšek et~al.(2015)Makovšek, Benezech, and Perkins]{makovsek2015efficiency}
Dejan Makovšek, Vincent Benezech, and Stephen Perkins.
\newblock Efficiency in railway operations and infrastructure management.
\newblock OECD/ITF Roundtable Reports 177, OECD Publishing, 2015.

\bibitem[McDaniels et~al.(2008)McDaniels, Chang, Cole, Mikawoz, and Longstaff]{mcdaniels2008fostering}
Timothy McDaniels, Stephanie Chang, Darren Cole, Joseph Mikawoz, and Holly Longstaff.
\newblock Fostering resilience to extreme events within infrastructure systems: characterizing decision contexts for mitigation and adaptation.
\newblock \emph{Global Environmental Change}, 18\penalty0 (2):\penalty0 310--318, 2008.

\bibitem[Wilbanks et~al.(2013)Wilbanks, Fernandez, Backus, Garcia, Jonietz, Kirshen, Savonis, Solecki, and Toole]{wilbanks2013climate}
Thomas Wilbanks, Steven Fernandez, George Backus, Pablo Garcia, Karl Jonietz, Paul Kirshen, Michael Savonis, Bill Solecki, and Leah Toole.
\newblock Climate change and infrastructure, urban systems, and vulnerabilities.
\newblock Technical report, U.S. Department of Energy, 2013.

\bibitem[Linkov et~al.(2014)Linkov, Bridges, Creutzig, Decker, Fox-Lent, Kr{\"o}ger, Lambert, Levermann, Montreuil, and Nathwani]{linkov2014changing}
Igor Linkov, Todd Bridges, Felix Creutzig, Jennifer Decker, Cate Fox-Lent, Wolfgang Kr{\"o}ger, James~H Lambert, Anders Levermann, Benoit Montreuil, and Jatin Nathwani.
\newblock Changing the resilience paradigm.
\newblock \emph{Nature Climate Change}, 4\penalty0 (6):\penalty0 407--409, 2014.

\bibitem[Abaza(2002)]{abaza2002optimum}
Khaled~A Abaza.
\newblock Optimum flexible pavement life-cycle analysis model.
\newblock \emph{Journal of Transportation Engineering}, 128\penalty0 (6):\penalty0 542--549, 2002.

\bibitem[Denysiuk et~al.(2017)Denysiuk, Moreira, Matos, Oliveira, and Santos]{denysiuk2017two}
Roman Denysiuk, Andr{\'e}~V Moreira, Jos{\'e}~C Matos, Joel~RM Oliveira, and Adriana Santos.
\newblock Two-stage multiobjective optimization of maintenance scheduling for pavements.
\newblock \emph{Journal of Infrastructure Systems}, 23\penalty0 (3), 2017.

\bibitem[Hussein et~al.(2017)Hussein, Gaber, Elyan, and Jayne]{hussein2017imitation}
Ahmed Hussein, Mohamed~Medhat Gaber, Eyad Elyan, and Chrisina Jayne.
\newblock Imitation learning: A survey of learning methods.
\newblock \emph{ACM Computing Surveys}, 50\penalty0 (2):\penalty0 1--35, 2017.

\bibitem[Sutton and Barto(2018)]{sutton2018reinforcement}
Richard~S Sutton and Andrew~G Barto.
\newblock \emph{Reinforcement Learning: An Introduction}.
\newblock MIT Press, 2018.

\bibitem[Andriotis and Papakonstantinou(2019)]{andriotis2019managing}
Charalampos~P Andriotis and Konstantinos~G Papakonstantinou.
\newblock Managing engineering systems with large state and action spaces through deep reinforcement learning.
\newblock \emph{Reliability Engineering and System Safety}, 191, 2019.

\bibitem[Yao et~al.(2020)Yao, Dong, Jiang, and Ni]{yao2020deep}
Linyi Yao, Qiao Dong, Jiwang Jiang, and Fujian Ni.
\newblock Deep reinforcement learning for long-term pavement maintenance planning.
\newblock \emph{Computer-Aided Civil and Infrastructure Engineering}, 35\penalty0 (11):\penalty0 1230--1245, 2020.

\bibitem[Silver et~al.(2016)Silver, Huang, Maddison, Guez, Sifre, Van Den~Driessche, Schrittwieser, Antonoglou, Panneershelvam, and Lanctot]{silver2016mastering}
David Silver, Aja Huang, Chris~J Maddison, Arthur Guez, Laurent Sifre, George Van Den~Driessche, Julian Schrittwieser, Ioannis Antonoglou, Veda Panneershelvam, and Marc Lanctot.
\newblock Mastering the game of {Go} with deep neural networks and tree search.
\newblock \emph{Nature}, 529\penalty0 (7587):\penalty0 484--489, 2016.

\bibitem[Vinyals et~al.(2019)Vinyals, Babuschkin, Czarnecki, Mathieu, Dudzik, Chung, Choi, Powell, Ewalds, and Georgiev]{vinyals2019grandmaster}
Oriol Vinyals, Igor Babuschkin, Wojciech~M Czarnecki, Micha{\"e}l Mathieu, Andrew Dudzik, Junyoung Chung, David~H Choi, Richard Powell, Timo Ewalds, and Petko Georgiev.
\newblock Grandmaster level in {StarCraft II} using multi-agent reinforcement learning.
\newblock \emph{Nature}, 575\penalty0 (7782):\penalty0 350--354, 2019.

\bibitem[Vora et~al.(2023)Vora, Thangeda, Grussing, and Ornik]{vora2023welfare}
Manav Vora, Pranay Thangeda, Michael~N Grussing, and Melkior Ornik.
\newblock Welfare maximization algorithm for solving budget-constrained multi-component {POMDPs}.
\newblock \emph{IEEE Control Systems Letters}, 7:\penalty0 1736--1741, 2023.

\bibitem[Thangeda and Ornik(2020)]{thangeda2020protrip}
Pranay Thangeda and Melkior Ornik.
\newblock Protrip: Probabilistic risk-aware optimal transit planner.
\newblock In \emph{International Conference on Intelligent Transportation Systems}, pages 1--6. IEEE, 2020.

\bibitem[Grussing et~al.(2006)Grussing, Uzarski, and Marrano]{grussing2006condition}
Michael~N Grussing, Donald~R Uzarski, and Lance~R Marrano.
\newblock Condition and reliability prediction models using the {W}eibull probability distribution.
\newblock In \emph{Applications of Advanced Technology in Transportation}, pages 19--24. American Society of Civil Engineers, 2006.

\bibitem[Kaelbling et~al.(1998)Kaelbling, Littman, and Cassandra]{kaelbling1998planning}
Leslie~Pack Kaelbling, Michael~L Littman, and Anthony~R Cassandra.
\newblock Planning and acting in partially observable stochastic domains.
\newblock \emph{Artificial Intelligence}, 101\penalty0 (1-2):\penalty0 99--134, 1998.

\bibitem[Watkins and Dayan(1992)]{watkins1992q}
Christopher~JCH Watkins and Peter Dayan.
\newblock Q-learning.
\newblock \emph{Machine Learning}, 8:\penalty0 279--292, 1992.

\bibitem[Williams(1992)]{williams1992simple}
Ronald~J Williams.
\newblock Simple statistical gradient-following algorithms for connectionist reinforcement learning.
\newblock \emph{Machine Learning}, 8:\penalty0 229--256, 1992.

\bibitem[Ng and Russell(2000)]{ng2000algorithms}
Andrew~Y. Ng and Stuart~J. Russell.
\newblock Algorithms for inverse reinforcement learning.
\newblock In \emph{International Conference on Machine Learning}, pages 663--670, 2000.

\bibitem[Zhao et~al.(2019)Zhao, Gaudoin, Doyen, and Xie]{zhao2019optimal}
Xiujie Zhao, Olivier Gaudoin, Laurent Doyen, and Min Xie.
\newblock Optimal inspection and replacement policy based on experimental degradation data with covariates.
\newblock \emph{IISE Transactions}, 51\penalty0 (3):\penalty0 322--336, 2019.

\bibitem[Huang et~al.(2010)Huang, Teng, and Lin]{huang2010budget}
Wen-Chih Huang, Junn-Yuan Teng, and Maw-Cherng Lin.
\newblock The budget allocation model of public infrastructure projects.
\newblock \emph{Journal of Marine Science and Technology}, 18\penalty0 (5):\penalty0 10, 2010.

\bibitem[Towers et~al.(2024)Towers, Kwiatkowski, Terry, Balis, De~Cola, Deleu, Goul{\~a}o, Kallinteris, Krimmel, and KG]{towers2024gymnasium}
Mark Towers, Ariel Kwiatkowski, Jordan Terry, John~U Balis, Gianluca De~Cola, Tristan Deleu, Manuel Goul{\~a}o, Andreas Kallinteris, Markus Krimmel, and Arjun KG.
\newblock Gymnasium: A standard interface for reinforcement learning environments.
\newblock \emph{arXiv preprint arXiv:2407.17032}, 2024.

\bibitem[Raffin et~al.(2021)Raffin, Hill, Gleave, Kanervisto, Ernestus, and Dormann]{stable-baselines3}
Antonin Raffin, Ashley Hill, Adam Gleave, Anssi Kanervisto, Maximilian Ernestus, and Noah Dormann.
\newblock Stable-baselines3: Reliable reinforcement learning implementations.
\newblock \emph{Journal of Machine Learning Research}, 22\penalty0 (268):\penalty0 1--8, 2021.
\newblock URL \url{http://jmlr.org/papers/v22/20-1364.html}.

\bibitem[Vora et~al.(2024)Vora, Grussing, and Ornik]{vora2024solving}
Manav Vora, Michael~N Grussing, and Melkior Ornik.
\newblock Solving truly massive budgeted monotonic pomdps with oracle-guided meta-reinforcement learning.
\newblock \emph{arXiv preprint arXiv:2408.07192}, 2024.

\bibitem[Blahoudek et~al.(2022)Blahoudek, Novotn{\`y}, Ornik, Thangeda, and Topcu]{blahoudek2022efficient}
Franti{\v{s}}ek Blahoudek, Petr Novotn{\`y}, Melkior Ornik, Pranay Thangeda, and Ufuk Topcu.
\newblock Efficient strategy synthesis for {MDPs} with resource constraints.
\newblock \emph{IEEE Transactions on Automatic Control}, 68\penalty0 (8):\penalty0 4586--4601, 2022.

\bibitem[Blahoudek et~al.(2021)Blahoudek, Cubuktepe, Novotn{\`y}, Ornik, Thangeda, and Topcu]{blahoudek2021fuel}
Franti{\v{s}}ek Blahoudek, Murat Cubuktepe, Petr Novotn{\`y}, Melkior Ornik, Pranay Thangeda, and Ufuk Topcu.
\newblock Fuel in {M}arkov decision processes ({FiMDP}): a practical approach to consumption.
\newblock In \emph{International Symposium on Formal Methods}, pages 640--656. Springer, 2021.

\bibitem[Zheng et~al.(2021)Zheng, Thangeda, Savas, and Ornik]{zheng2021optimal}
Wanzheng Zheng, Pranay Thangeda, Yagiz Savas, and Melkior Ornik.
\newblock Optimal routing in stochastic networks with reliability guarantees.
\newblock In \emph{International Conference on Intelligent Transportation Systems}, pages 3521--3526. IEEE, 2021.

\bibitem[Yazici et~al.(2017)Yazici, Ozguven, and Kocatepe]{yazici2017urban}
Anil Yazici, Eren~Erman Ozguven, and Ayberk Kocatepe.
\newblock Urban travel time variability in {N}ew {Y}ork {C}ity: A spatio-temporal analysis within congestion pricing context.
\newblock \emph{Transportation Research Board 96th Annual Meeting}, 2017.

\end{thebibliography}
\bibliographystyle{unsrtnat}

\end{document}